\documentclass[letterpaper, 10 pt, conference]{ieeeconf}  

\IEEEoverridecommandlockouts                              

\usepackage{graphicx} 
\usepackage{amsmath} 
\usepackage{amssymb}  
\usepackage{pifont}   
\usepackage{makecell} 

\title{\LARGE \bf
$A^3RNN$: Bi-directional Fusion of Bottom-up and Top-down Process \\
for Developmental Visual Attention in Robots
}

\author{Hyogo Hiruma$^{1, 2}$, Hiroshi Ito$^{1, 2}$, Hiroki Mori$^{1}$, and Tetsuya Ogata$^{1, 3}$
\thanks{*This work was supported by JST [Moonshot R\&D][Grant Number JPMJMS2031].}
\thanks{$^{1}$Hyogo Hiruma, Hiroshi Ito, Hiroki Mori, and Tetsuya Ogata are with the Department of
              Intermedia Art and Science, Waseda University, Tokyo, Japan
        hiruma.pyuma@ruri.waseda.jp, hiroshi.ito.ws@hitachi.com, ogata@waseda.jp,
        $^{2}$Hyogo Hiruma and Hiroshi Ito are with Research Development Group, Hitachi, Ltd. Tokyo, Japan,
        $^{3}$Tetsuya Ogata is with the National Institute of Advanced Industrial Science and Technology (AIST), Tokyo, Japan}
}

\begin{document}

\maketitle
\thispagestyle{empty}
\pagestyle{empty}

\begin{abstract}
This study investigates the developmental interaction between top-down (TD) and bottom-up (BU) visual attention in robotic learning. Our goal is to understand how structured, human-like attentional behavior emerges through the mutual adaptation of TD and BU mechanisms over time. To this end, we propose a novel attention model $A^3 RNN$ that integrates predictive TD signals and saliency-based BU cues through a bi-directional attention architecture.

We evaluate our model in robotic manipulation tasks using imitation learning. Experimental results show that attention behaviors evolve throughout training, from saliency-driven exploration to prediction-driven direction. Initially, BU attention highlights visually salient regions, which guide TD processes, while as learning progresses, TD attention stabilizes and begins to reshape what is perceived as salient. This trajectory reflects principles from cognitive science and the free-energy framework, suggesting the importance of self-organizing attention through interaction between perception and internal prediction. Although not explicitly optimized for stability, our model exhibits more coherent and interpretable attention patterns than baselines, supporting the idea that developmental mechanisms contribute to robust attention formation.
\end{abstract}

\section{INTRODUCTION}

The human cognitive system is fundamentally shaped by its ability to selectively attend to relevant information in complex environments. Attention allows humans to filter overwhelming sensory inputs, prioritize meaningful stimuli, and guide behavior effectively. Recent advances in cognitive science suggest that attention is not only a mechanism for selecting sensory input, but an active process driven by predictive inference. The \textit{free-energy principle} provides a unified theoretical framework to understand how the brain minimizes uncertainty by continuously refining internal representation about the environment \cite{friston2010free,friston2017active}. Under this framework, perception and action serve to reduce \textit{prediction errors}, ensuring that the brain’s expectations align with actual sensory perceptions. In free-energy principle, attention is proposed to regulate the \textit{precision} of these predictions, dynamically adjusting how much weight is given to certain sensory signals \cite{feldman2010attention,clark2013whatever}. The errors of high-precision signals are prioritized, whereas unreliable information is suppressed, optimizing cognitive and perceptual efficiency.

From a developmental perspective, infants refine their perceptual models by engaging in active exploration. Their attentional shifts are not random, but structured to maximize learning opportunities by minimizing prediction errors \cite{smout2019attention}. Studies on infant cognition suggest that attention is intrinsically goal-directed, seeking novel but predictable patterns that facilitate category learning and environmental understanding \cite{kidd2012goldilocks}. This dynamic interplay between sensory input and internal models provides a powerful framework for understanding not only biological development but also provides the basis to design effective and efficient models for intelligent robotics.

Human visual attention consists of two types: bottom-up (BU) and top-down (TD) attention. BU attention selects stimuli based on low-level visual features such as color, motion, or contrast \cite{itti2002model}, enabling fast and reactive perception. In contrast, TD attention is guided by internal states or task objectives, allowing an agent to selectively attend to goal-relevant regions \cite{desimone1995neural}. In human perception, these two modes operate in close coordination, enabling flexible adaptation to changing environments. Importantly, human attention emerges gradually through developmental processes. Infants learn to shift and sustain attention based on both sensory salience and interactive context, refining their allocation strategies through repeated experience \cite{johnson2010infants,oakes2023cascading}.

Recent robotic vision systems have begun incorporating attention mechanisms to enhance perceptual efficiency and task performance \cite{levine2016end}. Many of these models utilize either BU attention based on visual saliency \cite{ichiwara2021spatial}, or TD attention driven by task goals or prior knowledge \cite{seita2021learning}. However, most existing robotic attention models fail to capture the dynamic interplay between BU and TD attention, as they typically treat them as independent components. Even in approaches that integrate both cues \cite{hiruma2022deep}, the combination scheme is not clearly defined, making it difficult to capture how their relationship evolve over time through learning. As a result, these systems cannot account for the developmental nature of attention, limiting our ability to understand or replicate the flexible and adaptive characteristics of human-like attention.

To bridge this gap, we propose a novel approach \textbf{Amalgamated Active Attention ($A^3$)} RNN that integrates deep predictive learning \cite{suzuki2023deep} with a developmental framework for visual attention. Our model is based on the concept of \textit{precision control} from the free-energy principle, where the robot dynamically adjusts attention based on the reliability of sensory predictions. Specifically, we extend a pixel-based visual attention model Active Attention ($A^2$) RNN \cite{hiruma2022deep} with a Transformer-based self-attention architecture to explicitly leverage the BU signals to construct the TD attention. This framework allows attention to be driven not only by immediate sensory cues but also by learned expectations and ongoing predictive inference.

The contributions of this work are as follows:
\begin{itemize}
    \item We introduce a novel attention model that integrates BU and TD signals, enabling dynamic adaptation of attention allocation.
    \item We implement this model in a robotic learning environment to examine how BU and TD interactions evolve over time, providing insights into developmental attention mechanisms.
    \item Through empirical evaluation, we demonstrate that incorporating predictive inference improves attention stability and task performance in robot task learning.
\end{itemize}

By leveraging principles from cognitive science and developmental robotics, our approach offers a new perspective on how artificial agents can develop human-like attention mechanisms. This study not only advances robotic vision systems but also contributes to a deeper understanding of attention as an emergent property of predictive learning.
\section{Related Research}

Various visual processing models have been proposed in the context of learning-based robotics. However, each type of model exhibits different inductive biases, which make their suitability for modeling cognitive development uneven. 

One class of methods includes task-specific visual processing models such as object detectors \cite{liu2023lightweight}, image segmentation \cite{minaee2021image}, 3D pose estimators \cite{zimmermann20183d}, and point cloud regressors \cite{chen2022pq}. These models are highly effective in parsing complex visual scenes into semantically meaningful components, but these architectures typically require explicit label annotations or loss functions. As a result, any attention-like behavior that emerges through learning reflects the designer's biases making it unsuitable for investigating developmental process of TD and BU attentions.

As a more general alternative, transformer-based models such as Vision Transformers \cite{dosovitskiy2020image} have been used in end-to-end (E2E) robot learning. E2E models do not require explicit label annotations, which enables to learn attentional structures directly from data, potentially reducing prior bias. Furthermore, transformer models are well-suited for modeling multimodal data and temporal dependencies \cite{buamanee2024bi, al2024proceedings}, allowing them to form context-sensitive TD attention signals. However, the process of  treating all input tokens uniformly do not explicitly distinguish between BU and TD components. This makes it challenging to analyze how such attention structures evolve developmentally or how they relate to human-like attention mechanisms.

In contrast, pixel-based attention models attempts to mimic human-like attention more directly, by explicitly predicting a subregion of visual features as attention targets \cite{levine2016end}. For example, Spatial Attention RNN (SA-RNN) model \cite{ichiwara2021spatial} trains to extract sparse attention points from convolutional feature maps in E2E motion learning, identifying task-relevant image regions in a BU manner. In contrast, the $A^2RNN$ model \cite{hiruma2022deep} extends this approach by incorporating TD mechanisms. $A^2RNN$ combines convolutional features (as BU signals) with feedback of latent representations from an RNN (as TD signals) to compute attention points. Specifically, a vector derived from the RNN’s internal state is used as a query, while the CNN image features serve as keys and values, in a key-query-value attention scheme \cite{vaswani2017attention}. This structure is designed to be analogous to human cognitive structure, where intentional TD attention is derived from the states of the working memory, which serves a similar role to RNNs. The attended point is determined based on the similarity between the query and each feature location, enabling dynamic attention shifts in response to changing task contexts.

While $A^2RNN$ enables both BU and TD attention to interact in a task-driven manner, it exhibits certain limitations. In particular, the model has been shown to suffer from instability during training, often converging to different local optima depending on the random seed value for model weight initialization. For instance, in object picking tasks, the attention may incorrectly focus on non-task-relevant regions such as the screen periphery or empty space. This behavior is caused because the CNN and the RNN components co-adapt to minimize the reconstruction loss; in doing so, they may converge on attention strategies that are easy to predict but suboptimal for task performance. This phenomenon mirrors the “dark room problem” in the context of the free-energy principle \cite{friston2012free}, where an agent may converge on a behavior that minimizes prediction errors without being functionally appropriate. As a result, $A^2RNN$ converged on stable but semantically meaningless attention strategies unless additional optimization pressure is introduced to bias the learning.

In this work, we address these challenges by proposing a framework that decouples the BU and TD components of attention during learning, while maintaining their functional interaction. By integrating predictive learning with mechanisms that allow for developmental adaptation, we aim to support the emergence of attention strategies that are both flexible and cognitively meaningful.
\begin{figure*}[tbp]
    \centering
    \includegraphics[width=0.9\linewidth]{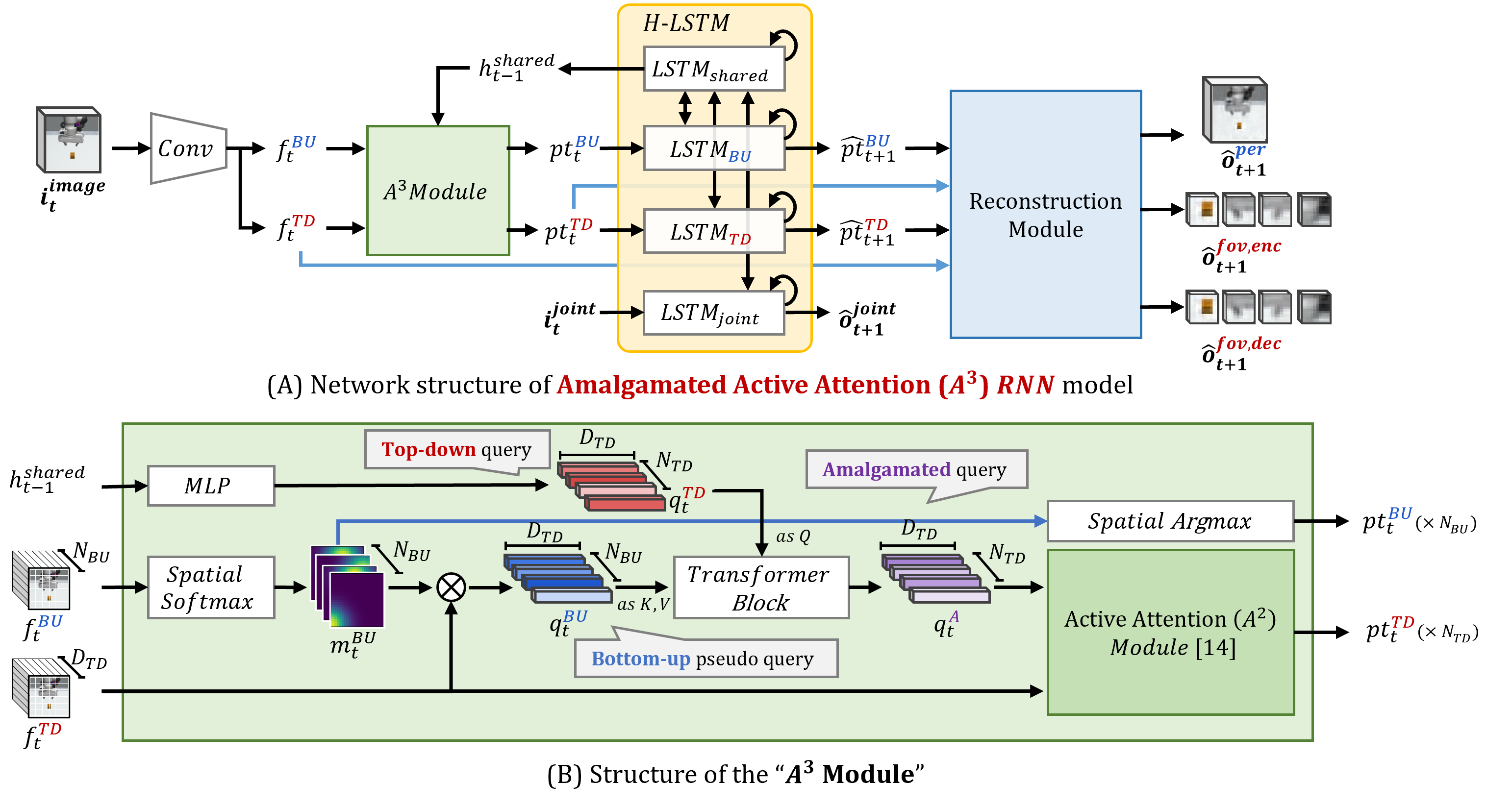}
    \caption{Proposed model structure. (A) The entire structure of the model, composed of $A^3$ module, Hierarchical LSTM (H-LSTM) module and Reconstruction module. (B) Detailed structure of $A^3$ module, where bottom-up queries and top-down queries are fused to an amalgamated query via a Transformer self-attention block.}
    \label{fig:model_structure}
\end{figure*}

\section{Method}
\label{sec:method}

\subsection{Overall Model Description}
The proposed $A^3RNN$ model extends our previous $A^2RNN$ model \cite{hiruma2022deep} in terms of prediction scheme of visual attention. Particularly, $A^3RNN$ introduces a fusion structure of BU and TD queries to enable the model to stably capture the targets (Fig.~\ref{fig:model_structure}(A)), whereas the previous $A^2RNN$ solely relied on TD queries that lack in training stability. 

The motion prediction scheme is based on a deep predictive learning framework \cite{suzuki2023deep}, which captures sensorimotor dynamics through the learning of robot motions and the resulting sensor observations: joint angle data ($i^{joint}$) and camera images ($i^{image}$). The predicted joint angle data is used as the command value for the next timestep, generating robot motions. $A^3RNN$ is composed of the (1) \textbf{Amalgamated Active Attention (A$^3$)} module, the (2) \textbf{Hierarchical Long-Short Term Memory (H-LSTM)} module, and the (3) \textbf{Reconstruction} module. Especially, the $A^3$ module plays a crucial role in stable attention learning, which enables to dynamically merge TD and BU visual attention information.

\subsection{$A^3$ Module}
The \textbf{A$^3$} module integrates BU and TD signals for adaptive attention allocation in sensorimotor tasks (Fig.~\ref{fig:model_structure}(B)). The module is designed to extract task-relevant image features by combining: (1) feedforward (BU) saliency from a convolutional neural network (CNN), and (2) feedback (TD) contextual signals from a recurrent model learning sensorimotor dynamics. Specifically, the BU information is derived from pixel-wise activation maps ($m_t^{BU}$) produced by a CNN, which reflect stimulus-driven saliency. The TD information is derived from the internal hidden state ($h_{t-1}^{shared}$) of an LSTM that learns the temporal structure of sensorimotor sequences. These signals are integrated via a Transformer-based encoder-decoder attention mechanism to generate context-aware attention query vectors ($q_t^A$).

The A$^3$ Module operates as follows:

\begin{enumerate}
    \item \textbf{Bottom-up attention map generation:}  
    Given a BU feature map $f_t^{BU} \in \mathbb{R}^{N_{BU} \times H \times W}$ from a CNN, spatial softmax is applied to each channel independently to obtain a list of normalized BU attention maps: $m_t^{BU} \in \mathbb{R}^{N_{BU} \times H \times W}$.

    \item \textbf{BU-derived pseudo-query vector extraction:}  
    Using each $m_t^{BU}$ as a spatial weighting mask, we compute the weighted average of a higher-level feature map $f_t^{TD} \in \mathbb{R}^{D_{TD} \times H \times W}$ to obtain a set of pseudo-query vectors: $q_t^{BU} \in \mathbb{R}^{N_{BU} \times D_{TD}}$.

    \item \textbf{TD query vector generation:}  
    The hidden state $h_{t-1}$ of the LSTM is transformed via a multi-layer perceptron (MLP) to produce TD query vectors: $q_t^{TD} \in \mathbb{R}^{N_{TD} \times D_{TD}}$.

    \item \textbf{Attention integration via Transformer:}  
    The BU pseudo-queries $q_t^{BU}$ are treated as keys and values, and the TD queries $q_t^{TD}$ serve as queries in a Transformer encoder-decoder structure. This yields integrated attention representations: $q_t^A \in \mathbb{R}^{N_{TD} \times D_{TD}}$.

    \item \textbf{TD attention point estimation:}  
    The integrated vector $q_t^A$ is used as the query in $A^2$ module \cite{hiruma2022deep} that estimates the final TD attention point $\mathrm{pt}_t^{TD}  \in \mathbb{R}^{N_{TD} \times 2}$ via pixel-wise similarity matching with CNN features $f_t^{TD}$.

    \item \textbf{BU attention point extraction:}  
    In parallel, spatial \texttt{argmax} is applied to each BU attention map $m_t^{BU}$ to obtain the corresponding bottom-up attention points $\mathrm{pt}_t^{BU} \in \mathbb{R}^{N_{BU} \times 2}$.
\end{enumerate}
$N_{BU}$ and $N_{TD}$ represents the number of predicted TD and BU attention points, respectively. $H$ and $W$ represents the height and width of the target image feature, $t$ represents the timestep, and $D_{TD}$ represents the channel dimension for TD attention query.

The A$^3$ module builds upon previous visual attention models, but extends them in two significant ways. First, it generates a set of pseudo-query vectors $q_t^{BU}$ from BU activations, enabling these to interact directly with TD queries through a Transformer structure. Second, by leveraging the attention mechanism of the Transformer, the model dynamically balances the influence of BU and TD pathways against the amalgamated attention, in a self-attention manner. 

This design allows the agent to modulate the relative weighting of prediction-driven (TD) and observation-driven (BU) signals depending on task demands and learning context. As such, it supports more flexible and adaptive attention formation than models with fixed attention strategies. In contrast to A$^2$RNN, w/hich embrace the risk of converging toward suboptimal local minima (e.g., focusing on low-motion areas to minimize loss), our model introduces an explicit decoupling and integration mechanism. This mitigates the tendency towards trivial prediction strategies and encourages the emergence of meaningful attention patterns in terms of task execution.

\begin{figure}[tbp]
    \centering
    \includegraphics[width=\linewidth]{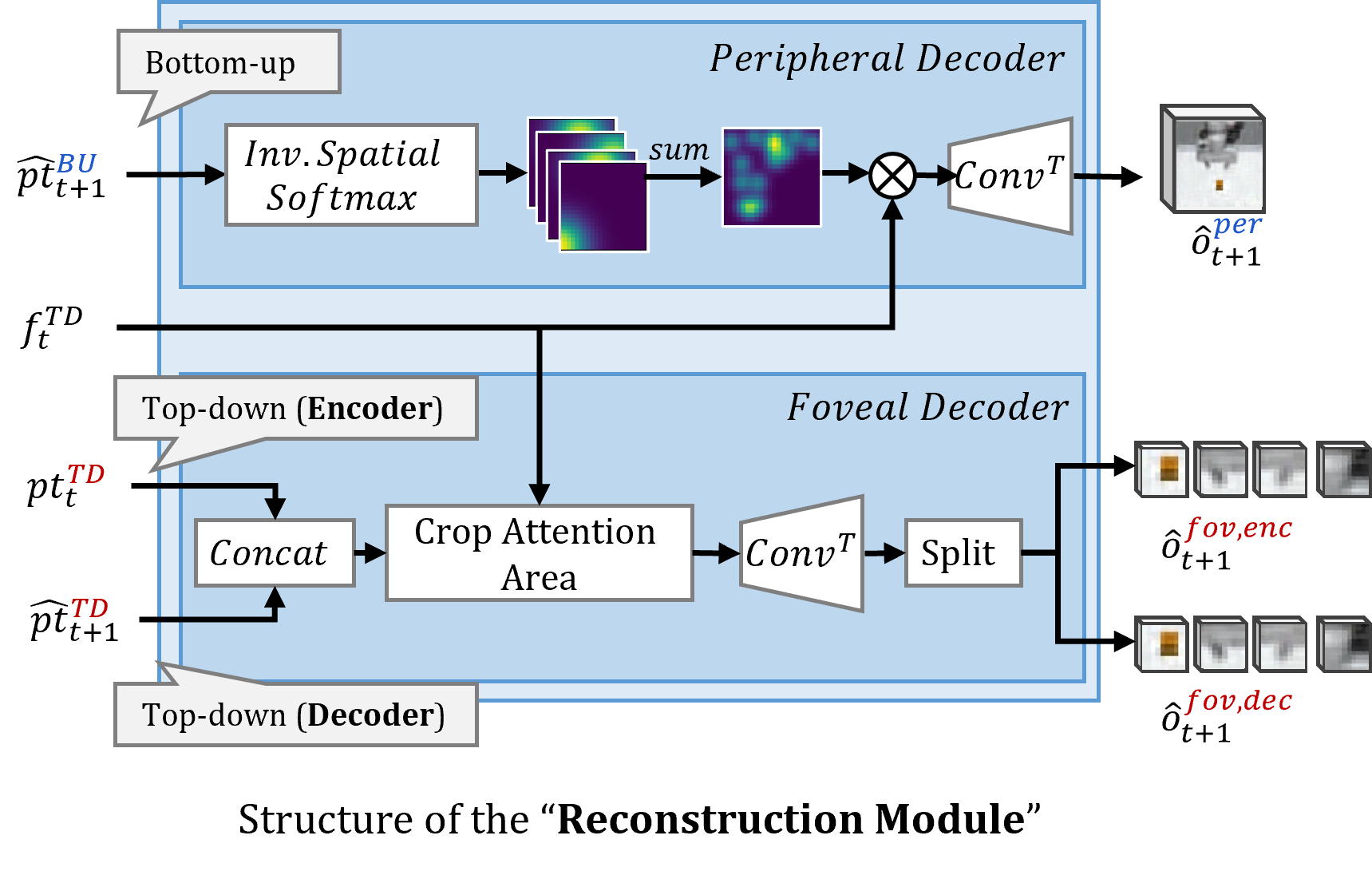}
    \caption{Detail structure of the reconstruction module. The predicted attention points of bottom-up and top-down attention are fed to reconstruct peripheral and foveal images, respectively.}
    \label{fig:recon_module}
\end{figure}

\subsection{Hierarchical LSTM Module}

The Hierarchical LSTM (H-LSTM) module consists of multiple LSTMs that are trained to capture dynamics from data of different modalities (image and joint angles) and temporal properties (slow and fast). The structure is inspired by the Multiple timescale RNN (MT-RNN) \cite{yamashita2008emergence} and the partitioning and unifying structure from $A^2RNN$ \cite{hiruma2022deep}.

The input is the attention point coordinates of TD ($pt_t^{TD}$) and BU ($pt_t^{BU}$), and the joint angles ($i_t^{joint})$. Each data is processed by independent LSTMs, which outputs $\widehat{pt}_t^{TD}$, $\widehat{pt}_t^{BU}$ and $\widehat{o}_t^{joint}$, respectively. Each LSTM mutually shares the information via shared LSTM ($\mathrm{LSTM}_{shared}$), which is a higher layer LSTM that integrates and redistributes the information to individual modality LSTMs.

\subsection{Reconstruction Module}
Fig.~\ref{fig:recon_module} shows the process of the Reconstruction module, which consists of a \textbf{Peripheral Decoder} and a \textbf{Foveal Decoder}. 

The reconstruction module predicts proximal future images for computing auxiliary losses that support visual attention acquisition. Inspired by the human visual system, it reconstructs two types of visual representations: a global low-resolution \textit{peripheral} image and a localized high-resolution \textit{foveal} image, corresponding to BU and TD attention, respectively.

In the \textbf{peripheral branch}, spatial coordinates from BU attention are converted into attention maps using inverse spatial softmax \cite{ichiwara2021spatial}. These maps are summed and applied as masks to CNN features ($f_t^{TD}$) that are passed via skip connections. The masked features are then decoded through a transposed convolution layers to reconstruct the entire image. This encourages the model to use BU attention to highlight salient regions and rely on the decoder to reconstruct the remaining background.

In the \textbf{foveal branch}, small 5$\times$5 feature patches centered at TD attention points are extracted from $f_t^{TD}$ (the process is noted as \textit{Crop Attention Area} in Fig.~\ref{fig:recon_module}). Both the current encoder-side attention $\mathbf{pt}^{TD}_t$ and the next-step decoder-side attention $\widehat{pt}^{TD}_{t+1}$ are fed for prediction. Each patch is decoded into a local image region using transposed convolution layers, resulting with multiple reconstruction patches at different attention regions.

Together, these reconstruction losses guide the model to form attention that are both saliency-aware and temporally coherent, improving the alignment between perception and sensorimotor dynamics.

\subsection{Training loss}

The proposed model is optimized by minimizing three types of losses: (1) a \textbf{body prediction loss} $L_{Body}$, (2) a \textbf{reconstruction loss} $L_{Rec}$, and (3) a set of \textbf{regularization losses} $L_{Reg}$. All losses are computed using full backpropagation through time (BPTT), and each aims to minimize the mean squared error (MSE) between predicted and target signals at each time step.

\begin{align*}
L 
&= L_{\mathrm{Body}} + \alpha\,L_{\mathrm{Rec}} + \beta\,L_{\mathrm{Reg}},\\
L_{\mathrm{Body}}
&= \mathrm{MSE}\bigl(\hat{o}^{\mathrm{joint}}_{0:T}\bigr),\\
L_{\mathrm{Rec}}
&= \mathrm{MSE}\bigl(\hat{o}^{\mathrm{per}}_{0:T}\bigr)
 + \mathrm{MSE}\bigl(\hat{o}^{\mathrm{fov,enc}}_{0:T}\bigr)
 + \mathrm{MSE}\bigl(\hat{o}^{\mathrm{fov,dec}}_{0:T}\bigr),\\
L_{\mathrm{Reg}}
&= \mathrm{MSE}\bigl(pt^{\mathrm{BU}}_{0:T},\,\hat{p}^{\mathrm{BU}}_{0:T}\bigr)
 + \mathrm{MSE}\bigl(\hat{o}^{\mathrm{fov,enc}}_{0:T},\,\hat{o}^{\mathrm{fov,dec}}_{0:T}\bigr)\\
&\quad + \Bigl(\mathrm{Dist}\bigl(pt^{\mathrm{TD}}_{1:T} - pt^{\mathrm{TD}}_{0:T-1}\bigr) - 0.1\Bigr)
 + \mathrm{Within}\bigl(\hat{p}^{\mathrm{TD}}_{0:T}\bigr).
\end{align*}
Note that MSE() represents the computation of MSE, where those with a single variable is compared against the ground truth. $Dist()$ indicates the computation of Euclidean distance and $Within()$ indicates the distance between the corresponding coordinates that are corrected to be within the image bounds. $T$ describes the length of the entire motion sequence.

\begin{figure*}[t]
   \centering
    \includegraphics[width=\linewidth]{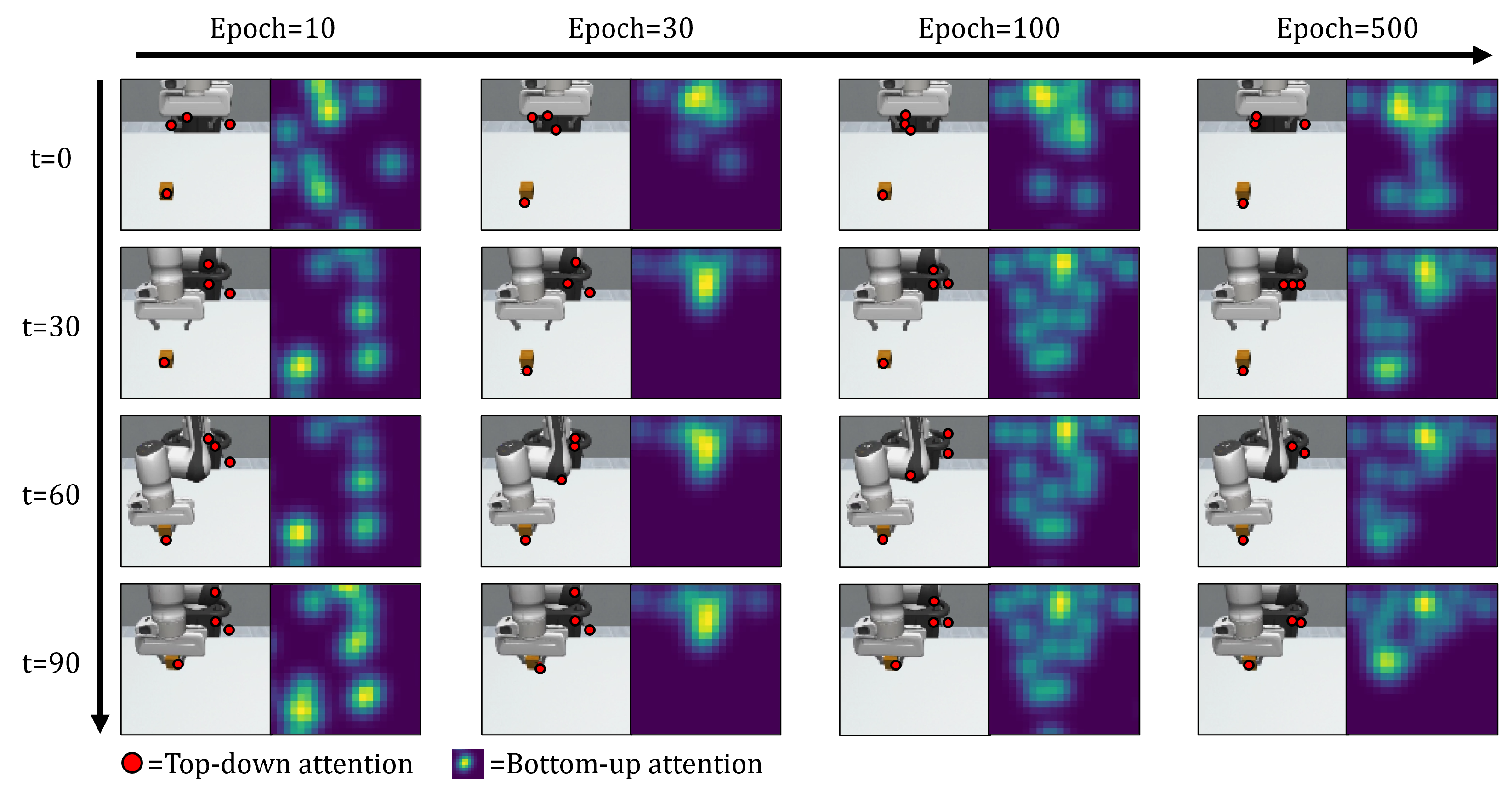}
    \caption{Comparison of the behaviors of top-down and bottom-up attention, visualized as red circles on the image and attention maps, respectively. Each column represents different training epochs and each row represents different timesteps. In the attention maps, brighter colors represent higher values which indicates higher importance score allocated to the pixels.}
    \label{fig:attention_dev}
\end{figure*}

\paragraph{Body Prediction Loss $L_{Body}$} 
This term supervises the predictive dynamics of the robot's motion, which measures the MSE between the predicted and ground truth joint angles during task execution.

\paragraph{Reconstruction Loss $L_{Rec}$}  
This loss is computed from the outputs of the reconstruction module and consists of three components:  
(1) the predicted peripheral image, (2) the reconstructed foveal image from the encoder-side TD attention point and (3) those from the decoder-side. By incorporating both current and future foveal predictions, this loss encourages temporal consistency in the TD attention trajectory while anchoring predictions to observable image data.

\paragraph{Regularization Losses $L_{Reg}$}  
To ensure meaningful and consistent attention behavior, we employ several regularization terms.  
In the first two terms, \textit{bi-directional consistency loss} \cite{hiruma2022deep} is applied between encoder and decoder attention. For BU attention, this is applied on the coordinates of the encoder and decoder attention points. For TD attention, however, we avoid directly penalizing coordinate discrepancies, since doing so empirically leads to degenerate solutions (e.g., the "dark room" problem). Instead, we compute the MSE between the corresponding foveal image reconstructions, encouraging perceptual consistency without enforcing identical coordinates. In the latter two terms, we add \textit{spatial validity constraints} to TD attention. One term penalizes attention points predicted outside the valid image bounds. Another term penalizes excessive displacement of TD attention between timesteps; specifically, a penalty is applied if the attention point moves over the length of 10\% of the image width or height. These constraints serve to stabilize attention behavior and prevent erratic attention predictions.

Together, these losses guide the model to form visual attention strategies that are not only predictive and goal-directed, but also consistent, interpretable, and grounded in visual structure.
\begin{figure}[t]
    \centering
    \includegraphics[width=0.5\linewidth]{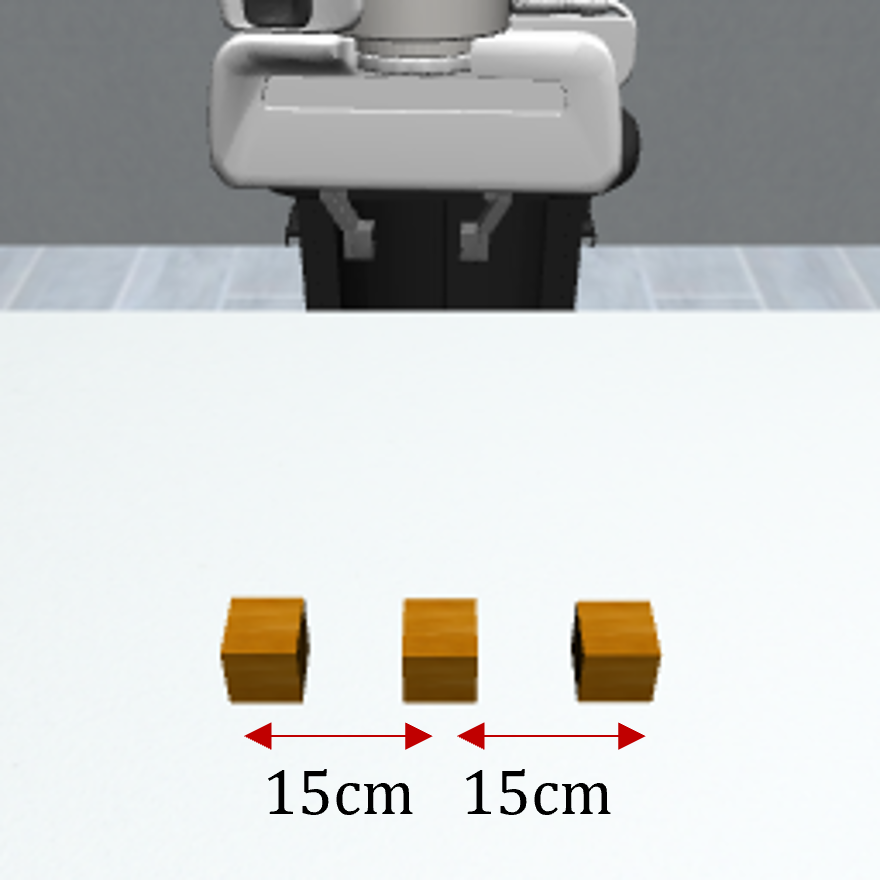}
    \caption{Setup of a simulator experiment. The robot arm was trained on a simple pick up task of a wooden box, placed at one of three different locations. }
    \label{fig:exp_setup}
\end{figure}

\section{Experiments}

To investigate the emergence and developmental dynamics of visual attention, we conducted a robot motion learning experiment in which a robot acquired sensorimotor behavior from a small number of demonstrations. We specifically analyzed the behaviors of the TD and BU attention points over the course of learning, focusing on how their interaction changes through it's experiences. The goal was to understand how the integration of TD and BU signals contributes to the formation and evolution of attention patterns, particularly under conditions where prior supervision is minimal and inductive biases are limited.

\subsection{Experimental Setup}

The experiment was conducted in a \texttt{robosuite} \cite{robosuite2020} simulator environment using a 7-DoF Panda robotic arm. Demonstration data were collected using a 3D mouse interface, allowing an operator to control the 6-DoF pose of the robot's end-effector in an intuitive manner. Each demonstration consisted of 120 timesteps, capturing both visual inputs and joint state information.

We evaluated the proposed model in a basic object-picking task. The task involved grasping a wood textured box that was positioned at one of three fixed locations on a tabletop (Fig.~\ref{fig:exp_setup}). The robot was trained to reach for the target object and pick it up. This setting provides a relatively simple and structured environment in which to examine how attention—particularly the relationship between BU saliency and TD task context—develops over the course of learning. We collected five demonstration sequences per box position, totaling 15 training sequences. The number of TD and BU attention points were set to $N_{TD}=4$ and $N_{BU}=16$, respectively.

This task setting enables clear observation of how attention behaviors evolve with training and how TD and BU attention begin to influence each other. Moreover, it allows us to evaluate whether the proposed model can acquire structured, goal-relevant attention patterns from limited data, in contrast to baseline models that often exhibit degenerate attention behaviors under similar low-supervision conditions.
\begin{figure}[tbp]
    \centering
    \includegraphics[width=\linewidth]{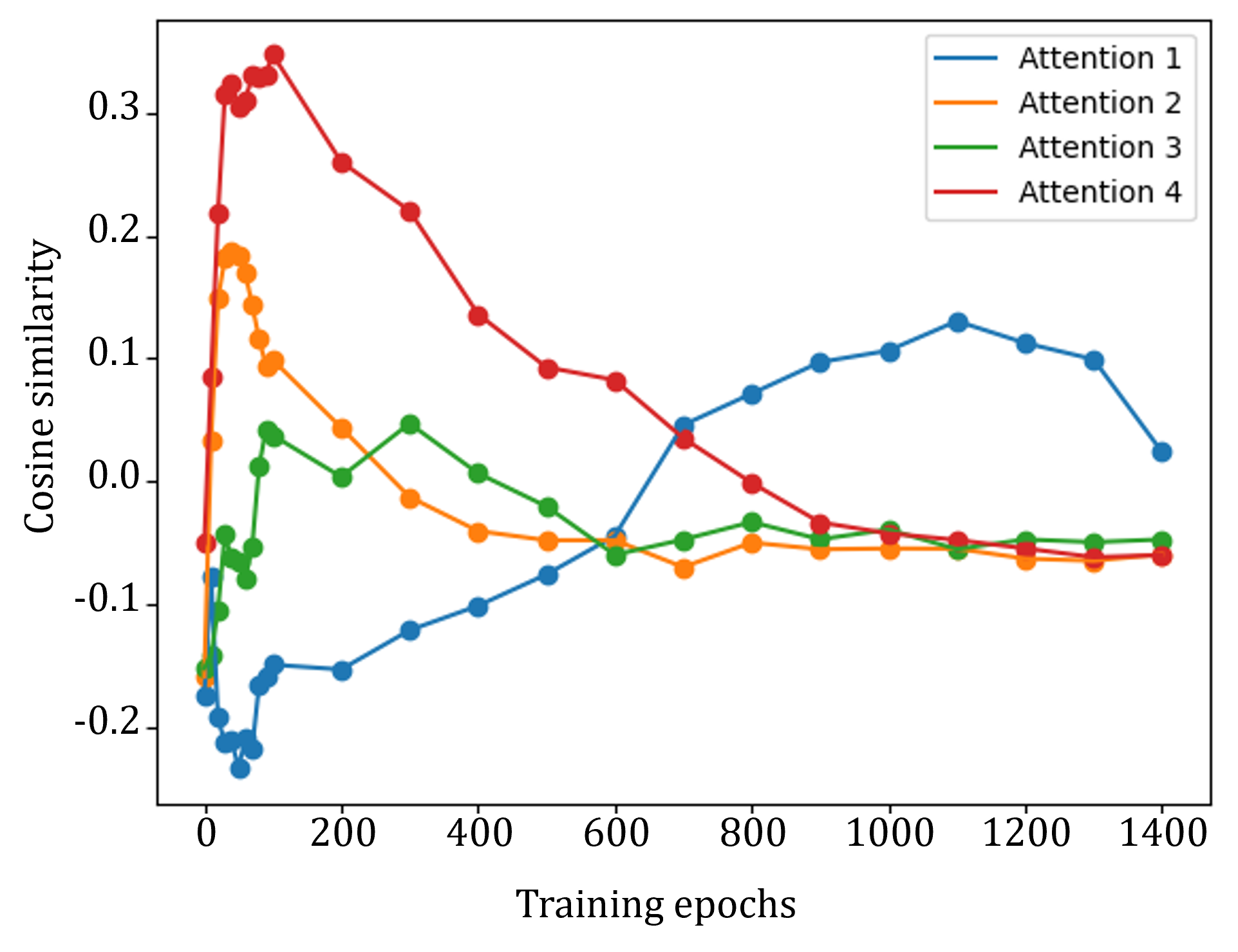}
    \caption{The developmental transition of average similarity between the amalgamated queries (Attention 1-4) and the BU pseudo queries.}
    \label{fig:query_similarity_dev}
\end{figure}

\section{Results}

\begin{table}[t]
    \centering
    \caption{Accuracy Comparison and Ablation Study Results}
    \begin{tabular}{c|c|c|c|c|c|c}
        \hline
        & \multicolumn{1}{c|}{Proposed} & \multicolumn{1}{c|}{A$^2$RNN} & \multicolumn{4}{c}{Ablation Study} \\ \cline{4-7} 
        & & & (1) & (2) & (3) & (4) \\ \hline
        Accuracy & \textbf{100\%} & 66.7\% & 8.3\% & 66.7\% & 91.6\% & \textbf{100\%} \\ \hline
    \end{tabular}
    \label{tab:accuracy_ablation}
\end{table}

\paragraph{Behavior Analysis of TD and BU Attention}
As shown in Fig.~\ref{fig:attention_dev}, the proposed model successfully acquired meaningful attention behavior during the object-picking task. The learned attention consistently focused on the target object throughout the entire action sequence and enabled reliable task completion. Compared to the A$^2$RNN model, which tended to fall into degenerate behaviors characteristic of the dark room problem, our model demonstrated more stable attention development while maintaining a biologically plausible structure (Table~\ref{tab:accuracy_ablation}).

To analyze the developmental dynamics of attention, we visualized the evolution of both TD and BU attention across training epochs. Fig.~\ref{fig:attention_dev} presents attention predictions generated from model snapshots at different training stages, sampled every 30 time steps. Each visualization includes the input camera image, the TD attention point (red circle), and the BU attention map. The TD attention point corresponds to the encoder-side prediction derived from the amalgamated query, which integrates both prediction-driven (TD) and stimulus-driven (BU) information. In contrast, the BU attention is visualized on the decoder side: predicted BU attention coordinates are transformed into image format and rendered as a single-channel saliency map. This map reflects regions prioritized for minimizing reconstruction error, thereby representing the model's implicit saliency estimation.

These visualizations reveal a notable shift in the interaction between TD and BU attention throughout training. In the early stage (e.g., epoch 10), BU attention is broadly distributed and appears almost random, though it often includes a few salient regions such as the object. TD attention, in contrast, initially follows BU's lead by focusing on areas to which BU attention is directed. This behavior stands in contrast to that of A$^2$RNN, where early TD attention fluctuates unpredictably across unrelated regions. This instability arises because A$^2$RNN directly derives TD queries from the RNN's hidden state at a point when temporal dynamics are not yet well learned. As a result, the model tends to focus on static, low-relevance regions, avoiding dynamic areas that ease the error minimization, despite being meaningless. Our results highlight the importance of BU pseudo-queries, which suppress initial randomness and effectively guide TD attention toward the correct target.

As training progresses, TD attention stabilizes around the object and the robot arm, while BU attention gradually shifts toward visually dynamic areas, such as the base of the arm. This behavior is likely due to BU attention being used for peripheral image reconstruction, requiring it to focus on task-relevant regions that benefit from skip-connected feature information, rather than relying solely on decoder biases. At intermediate stages (e.g., epoch 100), BU attention is broadly distributed across the task space—an effective policy for general reconstruction. However, by epoch 500, BU attention becomes more focused on the target object and the robot arm. As the figure shows, brighter areas in the BU map increasingly align with the TD attention point, indicating that BU attention is being shaped by TD influence for efficient parameter usage.

This mutual adjustment illustrates a bidirectional learning process: initially, BU attention selects salient stimuli that guide TD responses; later, TD attention becomes dominant and reshapes what BU interprets as salient. Over time, the model shifts from allocating attention across the full image to focusing on small, relevant regions—an efficiency-improving strategy for learning. Under the framework of the free-energy principle, this process corresponds to minimizing prediction error through both action (i.e., focusing attention) and perception (i.e., encoding saliency), suggesting the biological plausibility of the proposed model.

Fig.~\ref{fig:query_similarity_dev} further supports this claim, as the similarity between BU pseudo-queries and the amalgamated queries decreased over time. In early training epochs, the amalgamated queries strongly resemble BU pseudo-queries, showing high reliance to the BU features. Considering that all attention heads showed a peak in similarity (including the slow starting attention head 1), the existence of BU information can be thought as an important factor for the emergence of TD expectations. As the training progresses, each attention head diverged into independent latent representation, where the queries became increasingly orthogonal to the BU queries. This supports the qualitative analysis on Fig.~\ref{fig:attention_dev}, where TD and BU attention mutually affect to develop its attention. Notably, the model tended to increase BU dependence in one particular attention head (e.g., attention head 1), which exhibited relatively high prediction errors. This behavior resembles the context of predictive coding, where the unpredictable stimuli from the target induce BU processing \cite{rauss2013bottom}. Overall, these findings indicate that although attention initially depends on BU information, it gradually shifts toward being driven by the RNN’s internal state. This progression reflects the mutual influence of TD and BU attention mechanisms in developing coherent and context-sensitive visual attention.

\begin{table*}[t]
\centering
\renewcommand\theadfont{\normalsize\bfseries}
\caption{Model variants and their included components.}
\begin{tabular}{l|c|c|c|c|c|c|c}
\hline
\textbf{Model type} & \makecell{A$^2$\\Module} & \makecell{TD-BU\\integration} & \makecell{Peripheral\\recon.} & \makecell{Foveal\\recon.} & \makecell{Consistency\\reg. term} & \makecell{TD-BU\\integration (MLP)} & \makecell{Spatial\\reg. term} \\
\hline
A$^2$RNN            & \checkmark & \ding{55} & \ding{55} & \ding{55} & \ding{55} & \ding{55} & \ding{55} \\
Variant (1)       & \checkmark & \checkmark & \checkmark & \ding{55} & \ding{55} & \ding{55} & \ding{55} \\
Variant (2)       & \checkmark & \checkmark & \checkmark & \checkmark & \ding{55} & \ding{55} & \ding{55} \\
Variant (3)       & \checkmark & \checkmark & \checkmark & \checkmark & \checkmark & \ding{55} & \ding{55} \\
Variant (4)       & \checkmark & \ding{55} & \checkmark & \checkmark & \checkmark & \checkmark & \checkmark \\
A$^3$3RNN (Proposed) & \checkmark & \checkmark & \checkmark & \checkmark & \checkmark & \ding{55} & \checkmark \\
\hline
\end{tabular}
\label{tab:model_variants}
\end{table*}

\paragraph{Ablation Study}

We conducted an ablation study to evaluate the contribution of each component of the proposed model. Specifically, we tested the following four model variants (Table~\ref{tab:model_variants}). Each variant bases on the original A$^2$RNN, where individual components are added on per variant:
\begin{itemize}
    \item[(1)] Addition of TU and BU attention integration and BU peripheral reconstruction loss
    \item[(2)] Variant (1) with TD foveal reconstruction loss
    \item[(3)] Variant (2) with consistency regularization loss,
    \item[(4)] A$^3$RNN model, but with BU–TD query integration implemented via a Multi-Layer Perceptron (MLP) instead of a Transformer block.
\end{itemize}

The models were compared based on their success rates in directing attention to the correct object, evaluated across multiple runs with different random seeds. The results are summarized in Table~\ref{tab:accuracy_ablation}. Both the full proposed model and variant (4) achieved the highest stability, reaching a 100\% success rate across 12 trials. In contrast, variants (1), (2), and (3) showed lower success rates, indicating that each module contributes to improving the robustness of attention formation.

Although variant (4) reached comparable final accuracy to the full model, further analysis on the development of attentions revealed that it required nearly twice as many training epochs to capture task-relevant attentions, necessary for generating picking motions. Additionally, the interactive development of between BU and TD attention was less structured in this variant. , lacking the clear bi-directional influence observed in the Transformer-based model.

These results suggest that the dynamic integration of BU and TD queries via Transformer self-attention plays a critical role in learning efficiency. The Transformer mechanism enables the model to adaptively balance predictive top-down guidance with perceptual bottom-up saliency, resulting in more coherent, effective, and interpretable attention behaviors.
\section{Conclusion}

This paper investigated the developmental interaction between TD and BU visual attentions. Our objective was to explore how the two attentions co-evolve during the learning process, and how their mutual influence gives rise to structured, human-like attentional behavior. To this end, we proposed a novel visual attention model that integrates predictive TD signals and saliency-based BU cues using a bi-directional architecture for attention target selection. The proposed model showed high stability in acquiring task-relevant attentions, compared to existing models. Through analysis on robotic experiments, the behaviors of the acquired attentions resembled that of human cognitive behaviors. The emerged developmental strategy reflected the core ideas from the free-energy principle, where perception and action are jointly optimized to reduce prediction errors. Importantly, this mutually adaptive relationship emerged naturally through learning without the need for predefined supervision signals, leading to human-inspired developmental processes with more stable and interpretable attention behavior compared to baselines.

For future work, we plan to evaluate the model's predictability and robustness against complex and unstructured environments. Especially, training to stably select a target between identical objects remains a challenge due to attentional fluctuation between objects. We are also interested in extending the framework to incorporate other cognitive functions—such as reasoning under uncertainty or anticipatory control—that have traditionally been difficult to address due to the dark room problem. By grounding attention formation in developmental processes, we aim to better understand how embodied agents can acquire flexible, human-like cognition through sensorimotor learning.

\section*{ACKNOWLEDGMENT}
This work was supported by JST [Moonshot R\&D][Grant Number JPMJMS2031].

\bibliography{bibliography}
\bibliographystyle{unsrt}

\end{document}